# Fine-Tuning Models Comparisons on Garbage Classification for Recyclability

Umut Özkaya[1*] and Levent Seyfi [2]

[1]*Department of Electrical and Electronics Engineering/Konya Technical University, Konya, Turkey*
[2]*Department of Electrical and Electronics Engineering/Konya Technical University, Konya, Turkey*

*Corresponding author:* uozkaya@selcuk.edu.tr
[+]*Speaker:* uozkaya@selcuk.edu.tr

*Abstract –* In this study, it is aimed to develop a deep learning application which detects types of garbage into trash in order to provide recyclability with vision system. Training and testing will be performed with image data consisting of several classes on different garbage types. The data set used during training and testing will be generated from original frames taken from garbage images. The data set used for deep learning structures has a total of 2527 images with 6 different classes. Half of these images in the data set were used for training process and remaining part were used for testing procedure. Also, transfer learning was used to obtain shorter training and test procedures with and higher accuracy. As fine-tuned models, Alexnet, VGG16, Googlenet and Resnet structures were carried. In order to test performance of classifiers, two different classifiers are used as Softmax and Support Vector Machines. 6 different type of trash images were correctly classified the highest accuracy with GoogleNet+SVM as 97.86%.

## I. INTRODUCTION

The place of recycling in modern society is very important. In recycling process for today, separation of waste with manpower must take place in order to make a series of large filters. People may be confused about how products they consume are considered to be garbage. Within the scope of this study, it was to develop an algorithm for classification of garbage. Our aim is to increase efficiency of waste processing facilities and to identify non-recyclable wastes because garbage separation process is very difficult to separate garbage with 100% accuracy. The proposed method will be designed not only for environmental benefits but also for saving time and manpower.

In the recycling facilities, TrashNet is used for supervised learning with various images of differentiated materials and categorizing these images. The images obtained using a white background are estimated using fine-tuning models and different classification algorithms.

In 2012, Alexnet is a type of Convolutional Neural Network (CNN) architecture which was ImageNet Challenge winner, launched a new era in image classification [1]. The architecture used in this contest has a simple structure that is not deep. The performance is extremely high. AlexNet's effective performance in the ImageNet competition with a high degree of difficulty has led many researchers to work on CNN structures in the solution of image classification problems. One of the recent work on recycling is the automatic trash bin by TechCrunch Disrupt Hackathon team to determine whether the garbage is suitable for recycling with the Raspberry pi and a camera module. The process of this project is only to classify whether or not garbage is suitable for recycling [2]. Another project related to recycling is to categorize garbages as a smartphone application by utilizing imaging method [3]. With the application carried out, it is to ensure that citizens are followed up for recycling garbage in the neighborhood. They obtained data set needed to be trained Alexnet model through Bing Image Search. The accuracy rate obtained after training phase is approximately 87.69%. Using pre-trained model, a faster and higher accuracy network training was conducted. In another study, it was ensured that metal scrap was recycled by using formal properties. In this context, physical properties of material planned to be recycled were utilized. A number of methods were used to determine chemical and mechanical properties of wastes [4]. Material classification was made by using the Flicker database [5]. SIFT, color, micro texture and outline shape features from image dataset were used with Bayesian classification. Yang et al. conducted a series of features from images by support vector machines and scratch CNN structures were made using classification of garbage. The images in data set consist of glass, paper, metal, plastic, cardboard and other rubbish. There are approximately 400-500 images in each class. In the data set, deformation and logos of materials in images are slightly different from other data sets. SIFT features were extracted from images and classified by CNN structure. Many fine-tuned models were used in this study. In addition, a comparative analysis was performed by changing the classifier part [6]. Cenk et al. have tried many CNN model as strach and fine-tuned model. Inception-Resnet model in scracth models has achieved the highest classification success with 90% test accuracy. Fine-tuned models have achieved a 95% testing accuracy with DenseNet121 [7]. This paper is organized as four sections. Section I is introduction which consist of general information and literature works. Section II is materials and method parts.

Obtained results and discussion are in Section III. Finally, conclusion part is in Section IV.

## II. MATERIALS AND METHOD

### A. Deep Learning

Deep learning, called hierarchical learning, aims to analyse structure of data from simple to complex by using multilayer structures. In particular, effective features obtained by Convolutional Neural Network (CNN) make classification process much easier. In order to better understand convolutional neural networks, first layer on convolutional network identifies the simplest structures contained in the image, for example, through familiar two-dimensional images. Each of many filters in layer is obliged to find one of edges in a given image. This means that when an edge information contained in the image is taken into account, different filters serve for different angles of an edge, or for different shapes with other edges. The output of the first layer is feature maps that contain the information of structures that these filters detect. These outputs include information about various edge structures associated with the image. The next evolution layer reveals relationship edges that have been detected on the previous layer on features maps. Each convolution layer analyses correlations of combinatorial structures detected in previous layer with image. This flow, in which complexity increases with the number of layers, facilitates the acquisition of semantic information from structural information.

### B. Convolutional Neural Network

Convolutional neural networks (CNN) is a specialized state of multilayer neural network and is designed to detect geometric shape in image processing. In conventional multi-layer neural network, a neuron in first layer is connected with all neurons in next layer; convolutional layer establishes local connections on the output of previous layer. The fully connected layer performs a matrix multiplication. Convolutional layer uses convolution process, a linear mathematical process as in Eq 1:

$$y(t) = u(t) * x(t) = \int u(\tau) x(t-\tau) d\tau \qquad (1)$$

When Eq. 1 is used for continuous domain, Eq. 2 is obtained.

$$y[k] = \sum_j u[j] x[k-j] \qquad (2)$$

$x[k]$ can be written as $x[k-j]$ in Eq.2. Also, $x[k-j]$ can be transformed as $x[k+j]$ in Eq. 3. It is not effected the result.

$$y[k] = \sum_j u[j] x[k+j] \qquad (3)$$

In the convolution process performed with ANN, $w$ is a filter in convolution layer, $x$ is input of this layer and $f(.)$ is an activation function. The mathematical process performed by weight vector ($w$) of neurons on convolution layer on a spectral vector. Stride ($\zeta$) is a parameter, which is defined as the amount of shift on input window of filter window in convolution process, is also taken into account as in Eq. 4.

$$y_n[k] = f(w * x_n) = f(\sum_j w[j] x_n[\zeta k + j]) \qquad (4)$$

The reason for using $\zeta$ parameter, which is a positive integer, is to reduce size of large-sized inputs during the convolution by adding a sample sparse function to convolution process. In signal processing applications, length of signal obtained as a result of convolution of two signals of length $L_1$ and $L_2$ according to Eq. 2. The convolution process in CNN is different from signal processing; it calculates by taking into account only the places where short signal is over long one, and does not take into account partial overlaps in the regions where convolution starts and ends.

### C. SoftMax

SoftMax regression is used for last layer of a deep network where classification task is completed. The feature vector obtained by the previous layers is input of SoftMax regression layer. SoftMax regression is a generalized version of logistic regression and is used in multi-class classification problems. SoftMax regression expression is as in Eq. 5:

$$q_n = W^T \hat{x}_n + w_0 = [q_{n,1} \ q_{n,2} \ ..... \ q_{n,k}]^T \qquad (5)$$

SoftMax function is used to obtain uniform regression values between classes:

$$P(y_n = c | x_n) = \frac{\exp(q_{n,c})}{\sum_{j=1}^{k} \exp(q_{n,j})} \qquad (6)$$

In the result of softMax regression, class in which $x_n$ is assigned is given by Eq. 7.

### D. Support Vector Machine

Support Vector Machine (SVM) is an instructive technique that can be used for classification and regression [8]. It was developed by Vapnik based on statistical learning theory [9]. SVM is based on principle of processing features in a space that is much higher than original feature space. Data can always be divided into two different categories with a hyperplane [10]. SVM allows selection of plane that makes up the largest separation range from data. The aim is to choose right one to make classification with a minimum error when an unknown dataset is encountered. When margin is maximum, classification error is decreased.

$$<w, x_i> + b \geq 1 \quad \text{if} \quad (y_i = 1) \qquad (7)$$

$$<w, x_i> + b \geq -1 \quad \text{if} \quad (y_i = -1) \qquad (8)$$

Where $w$ is weighted vector, $b$ is bias value, $x$ is any point on a multiple plane and $y$ is label. Distance from $x_i$ is point to a hyper plane:

$$d(w, b; x_i) = \frac{|<w, x_i> + b|}{\|w\|} \qquad (9)$$

At the same time, margin between two hyperplane is represented as in Eq. 10:

$$\min_{x_i; y_i=1} d(w,b;x_i) + \min_{x_i; y_i=-1} d(w,b;x_i) \quad (10)$$

// *w* // should be minimized in order to maximize margin range [11]. The next stage involves solving quadratic optimization problem with linear constraints.

*E. Dataset*

The main purpose of this study is to realize recycling process independently of manpower. For this reason, it has been tried to create infrastructure of an intelligent system by benefiting from deep learning methods. We have used TrashNet dataset to train our proposed CNN network [12]. There are 6 classes in this data set such as glass, paper, cardboard, plastic, metal and garbage. The data set created using a white background consists of 512x384 size images. Fig. 1 shows some images belonging to data set.

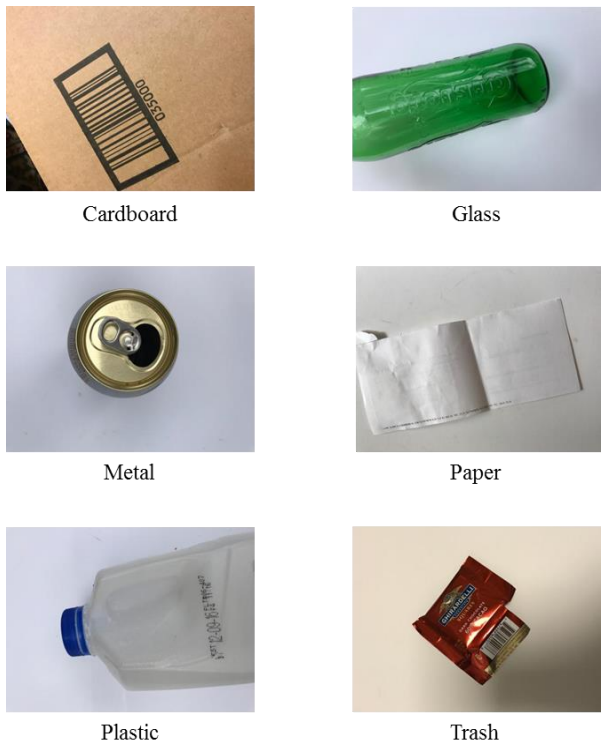

Fig. 2 Samples of Dataset

*F. Proposed Method*

Fine-tuned CNN structures were used in proposed method. These structures are AlexNet, VGG-16, GoogleNet, ResNet and SquezeeNet. Transfer learning is based on use of these structures. In the context of transfer learning, filter and weight values of previously trained CNN networks with millions of images are assigned as initial values in the training process to be performed for our own images. In this way, convergence of network has been realized more quickly and accurately. At the same time, Softmax and SVM were used in order to observe effect of classifiers on network performance as in Fig. 2.

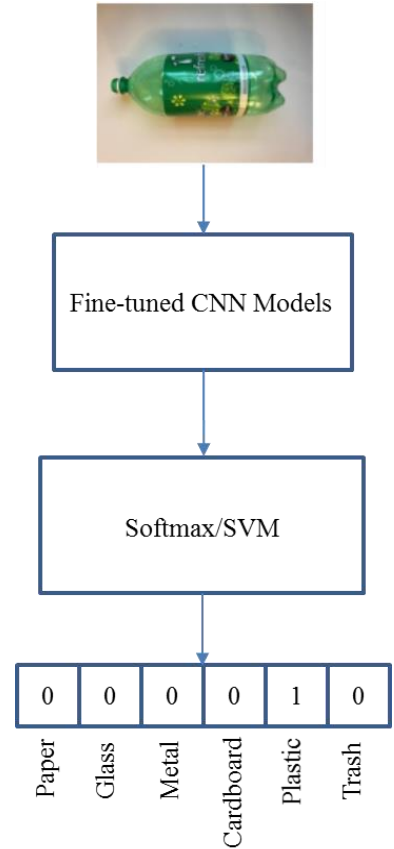

Fig. 2 Proposed Method

III. RESULTS AND DISCUSSION

We performed a comparative analysis for classification of images in TrashNet data set. In this analysis, we used half of data set for testing data without using any data augmentation method. In the context of our proposed method, AlexNet, GoogleNet, ResNet, VGG-16 and SquezeeNet were used as a fine tuned model. As can be seen in Table 1, we tried to obtain the highest classification accuracy by using two different classifiers such as Softmax and SVM. Table 2 and Table 3 provide classification accuracy of scratch and fine-tuned models in literature.

Table 1. Results of Proposed Methods

| Model | Accuracy of Fine-tuned Models (%) | | Data Aug. | Epoch |
|---|---|---|---|---|
| | **Softmax** | **SVM** | - | 200 |
| **AlexNet** | 87.14 | 97.23 | - | 200 |
| **GoogleNet** | 88.10 | **97.86** | - | 200 |
| **ResNet** | 89.38 | 94.22 | - | 200 |
| **VGG-16** | 90 | 97.46 | - | 200 |
| **SquezeeNet** | 80.43 | 90.17 | - | 200 |

In Table 1, accuracy of proposed method with Softmax is reached 90%. When classifier is assigned as SVM, classification accuracy goes up 97.86%. In our proposed method, Squezeenet + Softmax method is the lowest accuracy success with 83.43%. The highest accuracy rate belongs to GoogleNet + SVM with 97.86%.

Table 2. Results of Scratch Models [7]

| Model | Test Accuracy | Data Aug. | Epoch |
|---|---|---|---|
| ResNet | 75% | - | 100 |
| MobileNet | 76% | - | 500 |
| Inception ResNetV2 | 90% | + | 200 |
| DenseNet121 | 85% | - | 100 |
| DenseNet169 | 82% | + | 100 |
| DenseNet201 | 85% | - | 200 |
| Xception | 85% | + | 100 |

Table 3. Results of Fine-tuned models [7]

| Model | Test Accuracy | Data Aug. | Epoch |
|---|---|---|---|
| DenseNet121 | 95% | + | 200 |
| Inception ResNetV2 | 87% | + | 200 |
| RecycleNet | 81% | + | 200 |

When Table 2 is examined, classification results were obtained from scratch models with 90%. The classification accuracy with fine-tuned models in Table 3 is 95% despite data augmentation. The results in the literature are lower than results of our proposed method.

## IV. CONCLUSION

The classification of trash within the scope of recycling is possible with machine learning methods. Further data are needed to achieve higher accuracy rates. In the context of our proposed model, we have achieved high classification success without using any method of data augmentation. Studies show that the number of images and classes in the data set can be increased and a more comprehensive recycling project can be realized. In addition, we have achieved higher classification success with SVM classifier instead of Softmax. As a result, SVM is a more successful classifier than Softmax.